\pdfoutput=1
\documentclass{article}
\usepackage{iclr2027_conference,times}

\usepackage[T1]{fontenc}
\usepackage[utf8]{inputenc}
\usepackage{latexsym}
\usepackage{microtype}
\usepackage{graphicx}
\usepackage{booktabs}
\usepackage{amsmath,amsfonts}
\usepackage{nicefrac}
\usepackage{xcolor}
\usepackage{multirow,multicol}
\usepackage{enumitem}
\usepackage{xspace}
\usepackage{adjustbox}
\usepackage{tabularx}
\usepackage{ragged2e}
\usepackage{longtable}
\usepackage{hyperref}
\usepackage{url}
\hypersetup{hidelinks}
\usepackage{colortbl}

\newcommand{\our}{\textsc{Bookmarks}\xspace}
\title{\our: Checkpoint-Accelerated Active Grounding for Role-playing}

\author{Letian Peng, Ziche Liu, Yiming Huang, Longfei Yun,
Kun Zhou, Yupeng Hou, Jingbo Shang \\
University of California, San Diego \\
  \texttt{\{lepeng, jshang\}@ucsd.edu}\\
  }

\iclrfinalcopy 

\begin{document}
\maketitle
\lhead{Preprint.}
\begin{abstract}
    Memory systems are critical for role-playing agents (RPAs) to maintain long-horizon consistency. However, existing RPA memory methods (e.g., profiling) mainly rely on incremental summarization, whose compression discards details which become inaccessible to subsequent grounding. To address this issue, we propose a search-based memory framework called \textbf{\underline{\our}} for \textbf{active grounding}, which retains access to the full preceding storyline and collects task-relevant information on demand. Since summarizing the preceding storyline anew for each grounding request incurs substantial redundant computation, \our introduces \textbf{passive updating} to reuse earlier search results as checkpoints. Each \textbf{bookmark} represents the \textbf{content} about a particular aspect (\textbf{section}) of story information at a specific synchronization \textbf{point}. For current task, \our searches for only useful contents, reuses existing semantically equivalent bookmarks or initializes new ones, and synchronizes the selected ones from their stored checkpoints to the current scene. A reused bookmark thus only needs to process the newly observed storyline suffix, avoiding repeated synchronization. We evaluate \our across six narrative artifacts involving 47 characters and 9,537 test cases against non-active grounding baselines, covering next-action prediction and challenging, human-authored reasoning questions from a mystery game. \our improves next-action fidelity in a five-dimensional evaluation (emotion, intent, causality, position, and content) and raises mystery-game reasoning accuracy from 39.44\% to 45.42\% over the strongest baseline. Further analysis shows that (1) active grounding amplifies the benefits of additional sampling; (2) improvements span all five evaluation dimensions; and (3) reuse preserves quality at lower cost, while larger bookmark budgets yield continued gains.
\footnote{Code: \href{https://github.com/KomeijiForce/BOOKMARKS_Koishiday_2026}{KomeijiForce/BOOKMARKS\_Koishiday\_2026}}
    %
\end{abstract}

\section{Introduction}



Role-playing agents (RPAs)~\citep{survey_rp,chenpersona} require information about characters' evolving states and behaviors to remain faithful throughout a storyline. Existing memory methods either retrieve similar observations, limiting grounding to selected examples, or incrementally summarize observations into character profiles, leaving discarded details inaccessible. Introducing search-based grounding from general search agent~\citep{search-r1} retains access to the full preceding storyline, allowing RPAs to propose search targets based on the current scene and actively collect relevant evidence from earlier observations. A naive implementation is costly as each search processes the storyline from the beginning. Inspired by readers who leave \textbf{bookmarks} to revisit useful information, we propose \textbf{\our}, a memory framework that supports active grounding for RPAs and further accelerates via reusable checkpoints.

Specifically, \our imitates human readers by maintaining a pool of bookmarks inserted at different positions in the storyline. Each bookmark contains (1) a section title $q$; (2) the maintained content $y$; and (3) a synchronization point $p$ marking the latest processed storyline position. A bookmark $(q,y,p)$ represents section-specific grounding information at a specific time point.

\begin{figure}
    \centering
    \includegraphics[width=0.99\linewidth]{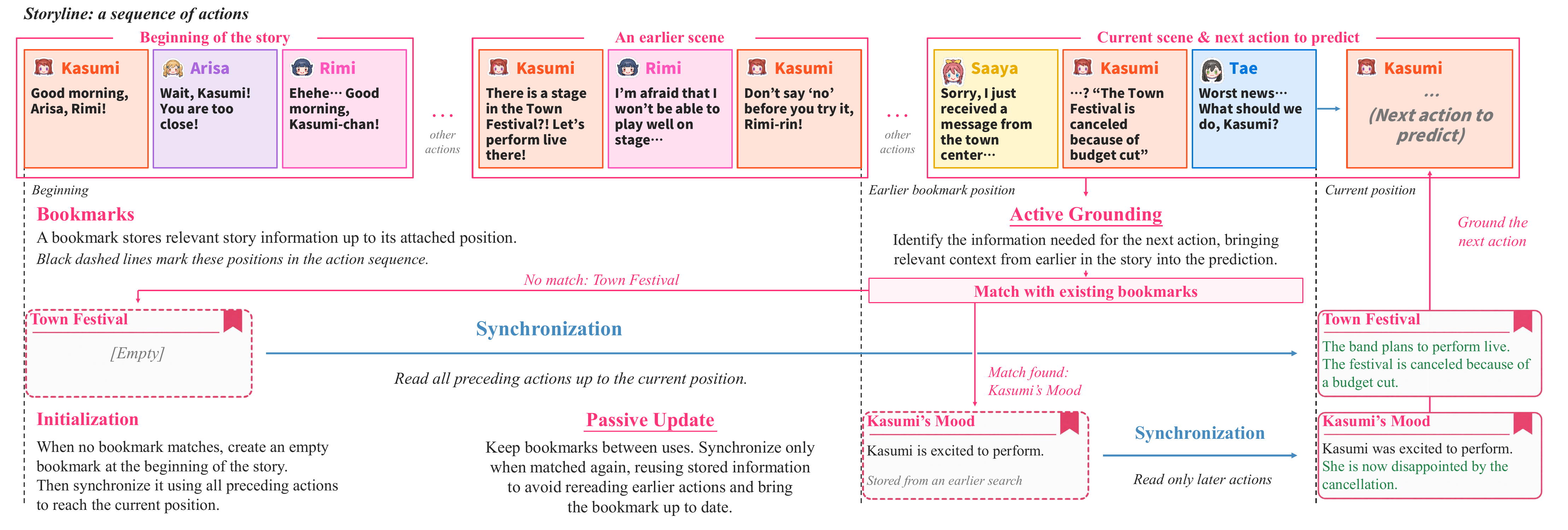}
    \vspace{-3mm}
    \caption{\our actively searches the preceding storyline for useful grounding information, while passive updating accelerates this search through checkpoint reuse.}
    \label{fig:intro_fig}
\end{figure}

Based on this data structure, \our grounds RPAs in three steps. (1) \textbf{Proposal} identifies sections that are potentially beneficial for grounding the current scene; (2) \textbf{Synchronization} reuses an existing semantically equivalent bookmark or initializes an empty one at the beginning of the storyline, then updates its content from the stored checkpoint to the current scene; (3) \textbf{Grounding} uses the synchronized contents to support RPA acting. We present a running example of \our in Figure~\ref{fig:intro_fig}, with more details in Figure~\ref{fig:bookmark_main}. In our implementation, each bookmark additionally has a type $\tau$ specifying how its content is represented and updated. Concept bookmarks maintain descriptions of entities or concepts. State bookmarks maintain ordered state nodes, refining the current node or appending a successor while retaining transition history. Behavioral bookmarks store actions together with their scenes and retrieve examples relevant to the current scene.

From a methodological view, conventional profile updating can be viewed as a special case of \our that maintains a single state bookmark through generic aggregation. In comparison, \our supports \textbf{active grounding} to seek information needed for the current scene. Its search cost is addressed by \textbf{passive updating}, which resumes selected bookmarks from their checkpoints and avoids repeatedly processing the same storyline prefix.

We evaluate \our on six diverse test artifacts drawn from an existing benchmark~\citep{peng2026deriving} and further extend the scope to cases made public after the evaluated model's (\texttt{gpt-5.1}) knowledge cutoff. We test each selected storyline from beginning to end with memory initialized empty, scoring next-action predictions by averaging binary match judgments across emotion, intent, causality, position, and content. We also evaluate accuracy on $284$ challenging human-authored reasoning questions with definite answers from ``Magical Girl Witch Trials'', made public after the same cutoff. These additions extend previous evaluations to newly released narratives and authentic reasoning, testing grounding beyond familiar storylines.

\our improves predictions across all five dimensions and raises reasoning accuracy from 39.44\% to 45.42\% over the strongest baseline. The largest average gains in content and causality suggest that storyline evidence helps ground what characters do and why, while generally larger gains when evaluating the best of three predictions indicate that this evidence can support stronger candidates with additional sampling. Checkpoint reuse maintains prediction quality comparable to searching from scratch while avoiding repeated updates. Increasing the bookmark budget also yields further gains, whereas enlarging Extract-and-Aggregate (ETA) profiles gives diminishing returns, supporting memory organization around current information needs over longer summaries.

Our contributions include (1) \our improves role-playing fidelity through active grounding over the full preceding storyline. (2) Passive updating accelerates this grounding through bookmark reuse while preserving prediction quality. (3) Expanded benchmarks and extensive analyses validate these benefits on post-cutoff narratives and challenging human-authored reasoning questions.

\section{Background and Related Work}

With the rapid development of LLMs' capability comes the ever-growing demand for more personalized interactions, to which role-playing agent (RPA) emerges as one of the central paradigms~\citep{chenpersona,tseng2024two}. These RPAs are expected to consistently produce in-character actions given story scene context, and the effective construction of such context is a \textit{grounding} problem. As storylines lengthen, a static context would fail to provide enough relevant information, and thus dynamic \textit{memory} systems become a key research direction. We accordingly cover \textbf{Evaluation}, \textbf{Training and Inference}, and \textbf{Memory} of role-playing agents.

\paragraph{Evaluation} of RPAs splits by granularity into holistic judgment and per-action scoring. \textit{Holistic judgment} aggregates over one of three character facets (identity, behavior, or knowledge) or bundles them. Identity is probed by psychiatric-style personality inventories~\citep{in_character, cheng2025psymem}, which rely on LLM-as-judge whose human alignment is known to be weak~\citep{zhoupersonaeval}. Behavior is tested in game, multi-agent, and social simulation testbeds~\citep{RPGBenchmark, zhousotopia, chen2024socialbench}. Knowledge is checked with factual or hallucination tests~\citep{shen2023roleeval, sadeq2024mitigating, ahn2024timechara}. Broad-coverage suites bundle several facets into one composite score~\citep{tu2024charactereval, lu2025rolemrc, he2025crab, ding2025rolermbench}. \textit{Per-action scoring}, in contrast, compares the next in-character action against a structured reference and is therefore facet-agnostic~\citep{codification}.

\paragraph{Training and Inference} are two routes to adapt a model over time. The training-time route bakes character into parameters through supervised fine-tuning on character experiences~\citep{shao2023character} or large-scale synthetic dialogue~\citep{moore2024rolellm, lu2024large, wang2025opencharacter, yang-etal-2025-crafting}, multi-character LoRA hot-swapping~\citep{yu2024neeko}, boundary- and personality-aware data~\citep{tang2024erabal, yang2025psyplay, ji2025enhancing}, and reinforcement-learning recipes~\citep{wang2025raiden, fang2025charm, liu2025cogdual}. These methods often suffer from plot scarcity, out-of-distribution hallucination, and an inability to absorb facts the storyline adds after training. The inference-time route leaves the backbone frozen and inserts structure between scene and response, like role-aware reasoning~\citep{tang2025thinking}, strategy-conditioned dialogue~\citep{SweetieChat}, retrieval-augmented exemplars~\citep{ricl}, and activation-level persona steering~\citep{chen2025persona}. Memory belongs to this same family but specifically deals with what content to store.

\paragraph{Memory} for RPAs focuses on what is stored (a compressed profile vs.\ an explicit structure) and how it's updated (statically before inference vs dynamically as scenes arrive). \textit{Static-profile methods} compress the storyline into a single profile re-attached every scene, in forms ranging from executable-function profiles~\citep{codification} to dialogue-recursive and topic-indexed summaries~\citep{wang2025recursively, zhong2024memorybank, lu2023memochat}. These methods guess what to keep and what to discard, and could lose critical cues that are useful later. \textit{Static-structure methods} fix a storage scheme ahead of time, like typed memory hierarchies~\citep{yan2023larp, sun2024identity}, event-and-relation graphs~\citep{bookworld, li2024graphreader, wang2025rolerag}, and mined if-then decision trees with distilled discriminators~\citep{peng2026deriving}. However, only a small, scene-dependent subset is needed at a time. \textit{Dynamic-profile retrieval} picks relevant profile entries for each scene~\citep{chen2025moom, huang2024emotional, wang2026memory}, but memory pool is fixed. Outside RP, state maintenance, world-model grounding, and long-form-story reasoning~\citep{yoneda2024statler, liu2024grounded, long_form_story_reasoning, yi2025score, xia2025storywriter} share one principle: keep enough state information to answer the next action-dependent question, and pull new information only when needed. 
\section{Our \our Framework}

\begin{figure}
    \centering
    \includegraphics[width=0.99\linewidth]{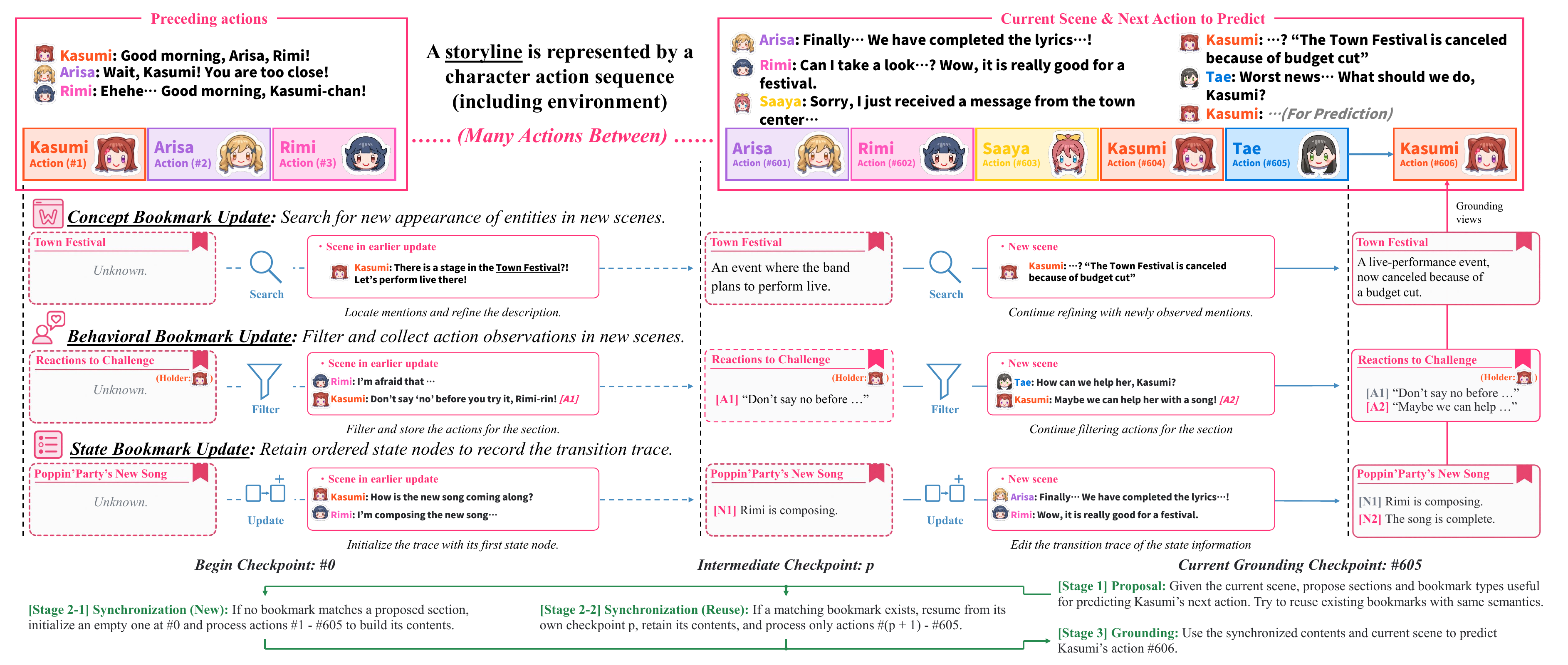}
    \vspace{-3mm}
    \caption{The role-playing grounding workflow of \our.}
    \label{fig:bookmark_main}
\end{figure}

\subsection{Preliminary}

\textbf{Storyline} can be viewed as a sequence of actions from different characters, including special ones such as \textit{``narration''} or \textit{``environment''}. We denote the storyline as $A = [a_1, a_2, \cdots, a_N]$, where $N=|A|$. A character sequence $C = [c_1, c_2, \cdots, c_N]$ assigns each $a_i$ to the character $c_i$ who takes it.

\paragraph{Role-playing Agents} RPAs aim to reproduce character behaviors by predicting the next action $a_i$ from the current scene $s_i$ and the target character $c_i$. In practice, $s_i$ is a local context before $a_i$, such as several preceding actions. Thus, an RPA can be viewed as $a_i \sim \mathrm{RPA}(\cdot \mid s_i, c_i)$.

\paragraph{Grounding Stage} Grounding augments the local scene with additional information before predicting $a_i$. This information may come from character profiles, retrieved examples, or storyline memory. We focus on data-driven grounding based on the preceding storyline $[a_1,\cdots,a_{i-1}]$.

\subsection{\our Framework}

Figure~\ref{fig:bookmark_main} illustrates the workflow of \our. Given a target character $c_i$ and scene $s_i$, \our constructs a grounding context from a memory bank $\mathcal{B}$ of reusable \textit{bookmarks}. Each bookmark maintains information under a \textit{section}, such as a character's current goals or responses to danger, at a specific point in the storyline.

At each prediction step, \our proposes useful sections, matches them with existing bookmarks or initializes new ones, and synchronizes the selected bookmarks to the current story point. It then composes their contents into a grounding context $g_i$ for prediction $a_i \sim \mathrm{RPA}(\cdot \mid s_i, c_i, g_i)$, which separates the choice of what to remember from information maintenance as the story unfolds.

\paragraph{Bookmark Data Structure}

A bookmark is represented as $b = (q, y, \tau, p)$,
where $q$ is a section title expressed as a noun phrase, $y$ is its maintained content, $\tau$ is the bookmark type, and $p$ is the synchronization point. The section identifies what is maintained, while its type determines how the content is represented.

We use three bookmark types, $\tau\in\{\texttt{concept, state, behavioral}\}$. \textbf{Concept} bookmarks maintain descriptions of named entities or concepts. \textbf{State} bookmarks maintain ordered traces of evolving states, such as locations, relationships, goals, or knowledge. \textbf{Behavioral} bookmarks store examples of a character's actions together with the scenes in which they occur. Each type thus preserves the information needed for grounding in a suitable form.

\subsection{Active Grounding}

The first stage selects what information would help ground the current prediction. Given $(s_i, c_i)$, the proposal module generates a small set of sections and their types:
\begin{equation*}
Q_i = \{(q_i^{(1)}, \tau_i^{(1)}), \cdots, (q_i^{(K)}, \tau_i^{(K)})\}.
\end{equation*}
Sections describe maintainable topics, such as \textit{``Alice's current goals''} or \textit{``Alice's responses to danger''}, so they can support later scenes as well as the current one.

This stage is active because the proposed sections depend on the current task. \our identifies the information relevant to the next action and uses it to select which parts of memory to access.

\subsection{Passive Updating}

After proposal, \our resolves each section to a bookmark and synchronizes the selected bookmarks. This stage is passive because a bookmark is updated only when selected for grounding.

\paragraph{Matching}

For each proposed section, \our searches for a bookmark with the same type, subject, and section meaning. \textbf{Reuse} activates a matching bookmark with its contents and synchronization point. \textbf{New} initializes an empty bookmark at $p=0$ when no matching section is found. Each selected bookmark is then synchronized from its own checkpoint.

\paragraph{Synchronization}

An activated bookmark is synchronized from its stored point $p$ to $i-1$ by processing the unseen suffix $[a_{p+1},\cdots,a_{i-1}]$. A type-specific update operator $U_{\tau}$ produces
\begin{equation*}
y' = U_{\tau}(q, y, [a_{p+1},\cdots,a_{i-1}], C),
\end{equation*}
yielding the updated bookmark $(q,y',\tau,i-1)$. The section stays fixed during type-specific updates.

\begin{itemize}[nosep,leftmargin=*]
    \item \textbf{Concept.} Synchronization locates mentions of the entity or concept in the unseen suffix and uses their local contexts to refine its description:
    \begin{equation*}
        y' = \operatorname{Refine}\bigl(q,y,\{(s_t,a_t):p<t<i,\;M(q,a_t)=1\}\bigr),
    \end{equation*}
    where $M(q,a_t)$ indicates whether $a_t$ mentions the entity or concept specified by $q$. The updated description accumulates information about how it is characterized in the story.

    \item \textbf{State.} For section $q$, the content $y$ records successive natural-language states:
    \begin{equation*}
        y = [y^{(1)},\cdots,y^{(L)}], \qquad L=|y|.
    \end{equation*}
    Each node includes its applicable time, place, or situation; the latest node $y^{(L)}$ describes the current state at $p$. The operator $U_{\texttt{state}}$ reads $[a_{p+1},\cdots,a_{i-1}]$ in chronological chunks. Each chunk preserves the trace if it adds no relevant information, refines the latest node within the same conditions, or establishes a successor state under changed conditions:
    \begin{equation*}
        y \leftarrow
        \begin{cases}
            y, & \text{keep},\\
            [y^{(1)},\cdots,y^{(L-1)},\tilde{y}], & \text{modify current},\\
            [y^{(1)},\cdots,y^{(L)},\tilde{y}], & \text{append state}.
        \end{cases}
    \end{equation*}
    Here $\tilde{y}$ is the state description written from $q$, the current trace, and the new chunk. An empty trace starts by appending its first supported state. Repeating this rule through $i-1$ yields $y'$, with earlier nodes retained as transition history.

    \item \textbf{Behavioral.} Synchronization filters scene--action examples by their relevance to the section and adds the retained examples to the stored set $y$:
    \begin{equation*}
        y' = y \cup \operatorname{Filter}\bigl(q,\{(s_t,a_t):p<t<i\}\bigr).
    \end{equation*}
    Keeping each action with its scene preserves the conditions under which the behavior occurred. At grounding time, a selection operator retrieves stored examples relevant to the current scene:
    \begin{equation*}
        \mathcal{E}_i(b) = \operatorname{Select}(s_i,y') \subseteq y'.
    \end{equation*}
    The grounding view is $\mathcal{E}_i(b)$, while $y'$ remains available for future updates and retrieval.
\end{itemize}

Across all types, synchronization advances the bookmark's checkpoint so later updates process only the newly observed suffix. Combined with active proposal, this limits maintenance to the memory selected for the current prediction, reducing repeated storyline processing during active grounding.

\paragraph{Grounding}

Finally, \our constructs $g_i$ from the synchronized bookmarks using a view suited to each type. Concept bookmarks contribute their descriptions. State bookmarks present their most recent nodes in chronological order through the current state, showing how the state evolved. Behavioral bookmarks contribute the selected scene--action examples $\mathcal{E}_i(b)$. Thus, the same behavioral section can provide different examples for different prediction contexts.

The same type-specific views are also applied to bookmarks with checkpoints near the current story point, providing recently maintained context. The resulting grounding context combines task-relevant concepts, evolving states, and situated behavioral examples to support the next action prediction.

\section{Benchmark}

\begin{table}
\caption{Examples of next-action prediction from Poppin'Party Band Story 0 (left) and human-authored multiple-choice questions from Magical Girl Witch Trials (right), with condensed context and English translation. The current scene during real benchmarking provides $10$ preceding actions.}
\label{tab:instance_case}
\centering
\small
\begin{adjustbox}{scale=0.8}
\begin{tabularx}{1.2375\linewidth}{@{}>{\raggedright\arraybackslash}X@{\hspace{1.2em}}>{\raggedright\arraybackslash}X@{}}
\toprule
\textbf{Next-Action Prediction} & \textbf{Multiple-Choice Questions} \\
\midrule
\textbf{Context}: Rimi had declined Kasumi's band invitation for fear of freezing onstage. When her sister's band was delayed, Kasumi and Arisa filled in, and Rimi joined them on bass.
&
\textbf{Context}: Each phone has a unique ID and can receive calls. Ema had called Miria. The recovered phone has Miria's distinctive case but no data; the handset inside could have been replaced. \\
\specialrule{0.4pt}{0.5pt}{0.5pt}
\textbf{Current scene (action sequence)}\par
\textbf{Kasumi:} A-ahaha, we got told off...\par
\textbf{Arisa:} Of course we did!\par
\textbf{Arisa:} And you dragged me into it! That was so embarrassing!\par
\textbf{Rimi:} But thanks to you two, I could get onstage... I was scared, but I had fun.
&
\textbf{Current scene (action sequence)}\par
\textbf{An-an:} That phone is Miria's! I never switched it!\par
\textbf{Nanoka:} But---\par
\textbf{An-an:} That's a baseless accusation! There is no evidence of a swap!\par
\textbf{Ema:} The evidence that the phone was switched... \\
\specialrule{0.4pt}{0.5pt}{0.5pt}
\textbf{Question:} What is Rimi's next action?
&
\textbf{Question:} Choose the correct continuation of Ema's statement. \\
\addlinespace[3pt]
\textbf{Reference action}\par
\textbf{Rimi:} If it's not too late, I'd like to be in a band with you two...!
&
\textbf{A.} No longer exists. \textbf{B.} Can be found at the crime scene.

\textbf{C.} Can be found by making a phone call. (\textbf{Answer:} C.) \\
\bottomrule
\end{tabularx}
\end{adjustbox}
\end{table}

\paragraph{Datasets} We construct eight test sets from a previous benchmark~\citep{peng2026deriving} and artifacts released after \texttt{gpt-5.1}'s knowledge cutoff. The collection covers \textit{The Melancholy of Haruhi Suzumiya} (Haruhi), \textit{JoJo's Bizarre Adventure} (JOJO), \textit{Fullmetal Alchemist} (FMA), \textit{A Game of Thrones} (AGOT), \textit{Poppin'Party} (PoPiPa), and \textit{Magical Girl Witch Trials} (Manosaba). These English and Japanese narratives span school comedy, supernatural adventure, political fantasy, music-centered daily life, and detective mystery. PoPiPa includes Band Story 0 and Event 321; the post-cutoff test cases include PoPiPa Event 321 and Manosaba. Table~\ref{tab:instance_case} illustrates the two evaluation tasks and Table~\ref{tab:main} summarizes dataset statistics. Appendix~\ref{apdx:character_info} provides artifact and character backgrounds.

\paragraph{Next-Action Prediction} We evaluate eligible character actions from beginning to end of each selected storyline, with memory initialized empty. For action $a_i$ of character $c_i$, the local scene $s_i$ contains up to $10$ preceding actions, and grounding uses the observed prefix $[a_1,\cdots,a_{i-1}]$. The original action is held out as the reference. We compare predictions with references across five dimensions: \textbf{emotion} (expressed emotion), \textbf{intent} (immediate purpose), \textbf{causality} (character reasoning), \textbf{position} (stance), and \textbf{content} (central proposition or action). Each dimension receives an independent yes/no judgment, scored as $100/0$. We report the mean match score across the five dimensions and all evaluated actions. We use \texttt{gpt-4.1}, whose reliability is supported by the previous benchmark~\citep{peng2026deriving}, as a judge distinct from all evaluated models. Our human validation of $200$ cases per dimension yields a mean agreement of $93.7\%$ with the judge, with a standard deviation of $1.36$ percentage points across dimensions.

\paragraph{Multiple-Choice Questions} We further include $284$ multiple-choice questions (MCQs) from Manosaba. These human-authored questions come from the original game, have definite correct answers, and require reasoning over accumulated story evidence. The questions became public in July 2025 after \texttt{gpt-5.1}'s knowledge cutoff, providing an additional test of reasoning on newly released material. The model selects an answer from the provided options, and we report accuracy against the original correct answers.

\section{Experiment}

\begin{table}
\caption{NAP reports the mean match score across five dimensions; MCQ reports accuracy (both in \%). Best@$k$ selects the best of $k$ predictions per action. MCQ uses Best@1 to avoid inflated accuracy from multiple guesses. Significance testing yields $p<0.05$.}
\label{tab:main}
\centering
\small
\begin{adjustbox}{max width=\linewidth,scale=0.9}
\begin{tabular}{llccccccccc}
\toprule
& \multirow{2}*{\textbf{Artifact}} & \multirow{2}*{\textbf{Haruhi}} & \multirow{2}*{\textbf{JOJO}} & \multirow{2}*{\textbf{FMA}} & \multirow{2}*{\textbf{AGOT}} & \multicolumn{2}{c}{\textbf{PoPiPa}} & \multicolumn{2}{c}{\textbf{Manosaba}} \\
& & & & & & BS0 & E321 & NAP & MCQ \\
\midrule
\multirow{5}*{\rotatebox{90}{\textbf{Info}}} & \textbf{Genre} & Sci-fi & Action & Fantasy & Politics & \multicolumn{2}{c}{Music} & \multicolumn{2}{c}{Mystery} \\
& \textbf{\#Character} & $5$ & $7$ & $5$ & $11$ & \multicolumn{2}{c}{$5$} & \multicolumn{2}{c}{$14$} \\
 & \textbf{\#Action} & $781$ & $1592$ & $1351$ & $1686$ & $1944$ & $306$ & $1593$ & $284$ \\
& \textbf{Released} & \multicolumn{5}
{>{\columncolor{gray!20}[4pt][4pt]}c}
{Before 20-01} & \multicolumn{1}
{>{\columncolor{gray!20}[4pt][4pt]}c}
{26-02} & \multicolumn{2}
{>{\columncolor{gray!20}[4pt][4pt]}c}
{25-07}  \\
\addlinespace[1pt]
& \textbf{Lang} & \multicolumn{4}
{>{\columncolor{gray!20}[4pt][4pt]}c}
{en} & \multicolumn{4}
{>{\columncolor{gray!20}[4pt][4pt]}c}
{ja}\\
\midrule
\multirow{5}*{\rotatebox{90}{\textbf{Best@1}}} & \textbf{Vanilla} & $32.21$ & $34.10$ & $34.21$ & $41.30$ & $36.05$ & $44.57$ & $38.10$ & $35.21$\\
& \textbf{RICL} & $32.86$ & $35.52$ & $35.28$ & $42.28$ & $37.50$ & $46.82$ & $39.82$ & $36.62$\\
& \textbf{ETA} & $33.47$ & $35.93$ & $35.72$ & $42.55$ & $38.65$ & $46.24$ & $40.22$ & $38.38$  \\
& \textbf{3-way ETA} & $34.17$ & $36.44$ & $36.47$ & $42.85$ & $38.25$ & $47.03$ & $40.84$ & $39.44$\\
& \textbf{\our} & $\textbf{36.37}$ & $\textbf{38.87}$ & $\textbf{37.88}$ & $\textbf{44.41}$ & $\textbf{39.87}$ & $\textbf{49.35}$ & $\textbf{43.97}$ & $\textbf{45.42}$\\
\midrule
\multirow{5}*{\rotatebox{90}{\textbf{Best@3}}} & \textbf{Vanilla} & $40.69$ & $45.57$ & $42.98$ & $50.31$ & $39.17$ & $48.27$ & $41.83$ & \multirow{5}{*}{\rotatebox{90}{\shortstack{Disabled\\by MCQ}}}\\
& \textbf{RICL} & $41.74$ & $47.83$ & $44.71$ & $53.75$ & $46.22$ & $53.21$ & $45.98$ & \\
& \textbf{ETA} & $42.16$ & $49.06$ & $45.38$ & $53.26$ & $44.82$ & $54.70$ & $46.65$ & \\
& \textbf{3-way ETA} & $42.52$ & $49.85$ & $46.84$ & $54.82$ & $45.89$ & $55.63$ & $48.19$ &\\
& \textbf{\our} & $\textbf{45.04}$ & $\textbf{52.21}$ & $\textbf{49.18}$ & $\textbf{56.84}$ & $\textbf{48.89}$ & $\textbf{59.23}$ & $\textbf{52.47}$ &\\
\bottomrule
\end{tabular}
\end{adjustbox}

\end{table}

\subsection{Baselines}

We compare \our with representative grounding methods that differ in how they access and maintain storyline information.

\begin{itemize}[nosep,leftmargin=*]
    
    \item \textbf{Vanilla} predicts character actions without additional grounding context, relying only on the parametric knowledge of the RPA.
    
    \item \textbf{Retrieval-based In-Context Learning}~\citep{ricl} \textbf{(RICL)} retrieves top-$k$ similar past scene-action pairs as grounding examples for the current scene.
    
    \item \textbf{Extract-and-Aggregate (ETA)}~\citep{coser} incrementally maintains a single character profile by summarizing new observations and aggregating them into the previous profile.
    
    \item \textbf{3-way ETA} divides the character profile into three fixed sections, \textit{experience}, \textit{state}, and \textit{behavior}. For each prediction, the RPA selects relevant sections and synchronizes only those sections through ETA-style recurrent updating.
    
\end{itemize}

\paragraph{Methodological Comparison} ETA can be viewed as a special case of \our since its character profile corresponds to a single state bookmark maintained through generic aggregation. 3-way ETA forms an intermediate setting by introducing selection among three fixed sections while retaining ETA-style updates. \our further allows sections to be proposed as needed and their contents to be maintained through type-specific representations and update rules. This progression lets us evaluate how flexible memory organization affect grounding quality.

\paragraph{Implementation}
We primarily use \texttt{gpt-5.1} as the RPA backbone and additionally evaluate open-source DeepSeek and Qwen models to assess generalizability across model families. RICL retrieves eight examples for ICL by scene semantic similarity using \texttt{qwen3-embedding-8b}. To match \our's additional output-token usage per synchronization round, ETA and 3-way ETA are prompted to limit their outputs to 1,024 tokens and 256 tokens per section, respectively. \our proposes at most three bookmarks per round, selects eight behavioral action examples using BM25, and updates state bookmarks in blocks of 64 actions. Prompts are in Appendix~\ref{apdx:prompts}.

\subsection{Main Results}

Table~\ref{tab:main} shows that \our achieves the best performance across diverse storylines and both evaluation tasks, supporting active grounding that maintains storyline memory around the needs of the current scene. All methods' gains over Vanilla highlight the value of explicit storyline evidence, while the advantage over RICL supports maintaining synchronized memory beyond retrieving individual examples. The general improvement from ETA to 3-way ETA suggests that selecting and updating relevant memory sections already benefits grounding. \our further improves on both by allowing flexible sections with type-specific representations and updates. Its generally larger gains under Best@3 suggest that this targeted grounding also helps the model produce stronger candidate actions when multiple predictions are sampled. These gains extend to post-cutoff narratives, supporting effective grounding on uncontaminated material, a setting rarely tested in prior work. The particularly large improvement on human-authored Manosaba MCQs further suggests that active grounding is especially suited to reasoning, where each question creates an immediate need to seek specific evidence from the observed storyline, which is further validated by cases in Appendix~\ref{apdx:case_studies}.

\begin{figure}[htbp]
    \centering
    \includegraphics[width=0.8\linewidth]{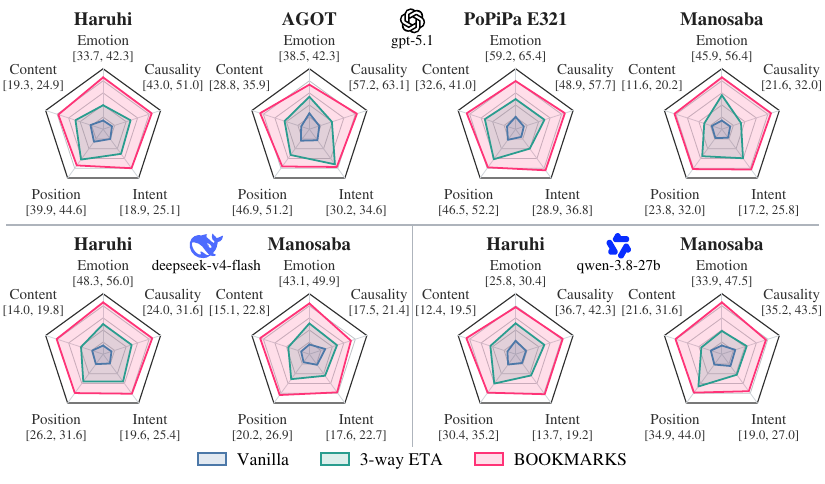}
    \vspace{-5mm}
    \caption{NAP match scores under Best@1. Bracketed intervals at each vertex give the axis limits (\%) at the radar center and outer boundary, respectively, with linear scaling between them.}
    \label{fig:multidimension_results}
\end{figure}

\paragraph{Multi-dimension Results} Figure~\ref{fig:multidimension_results} shows that \our achieves prominent gains over all five dimensions, yielding broad improvements in role-playing fidelity. For \texttt{gpt-5.1}, the largest average gains across storylines occur in \textit{Content} and \textit{Causality}. This pattern aligns with the goal of \our since the framework provides relevant storyline evidence to ground what a character does or says and the reasoning behind it. Maintaining synchronized concepts, evolving states, and situated behavioral examples supplies context that connects the next action to the character's preceding experiences and current circumstances.

\paragraph{Cross-Model Performance} The benefits also transfer to the open-source \texttt{deepseek} and \texttt{qwen} backbones as both show substantial improvements spanning all five dimensions. This broader pattern suggests that synchronized storyline memory supports multiple aspects of character behavior for backbones with more limited capabilities. Notably, \texttt{qwen-3.8-27b} benefits consistently from \our on both storylines, demonstrating that even a relatively small 27B-parameter model can make effective use of the framework.

\subsection{Efficiency Analysis}

\begin{figure}[htbp]
    \centering
    \includegraphics[width=\linewidth]{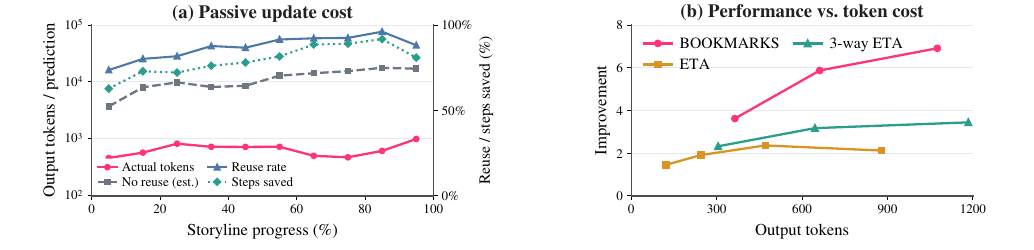}
    \vspace{-5mm}
    \caption{Efficiency analysis on Manosaba\protect\footnotemark. (a) Actual and estimated no-reuse output-token costs, with reuse rates and fractions of update steps saved on the right axis, averaged over 10\% storyline-progress buckets. (b) Best@1 NAP improvement over Vanilla versus output tokens.}
    \label{fig:grounding_efficiency}
\end{figure}
\footnotetext{We use Manosaba for analysis because its characters were unseen before the evaluated models' knowledge cutoffs, providing a clean setting without prior character knowledge.}

\our reduces the cost of maintaining storyline knowledge while continuing to benefit from a larger grounding budget. Figure~\ref{fig:grounding_efficiency} (left) shows that, as the storyline grows, reusing bookmarks avoids most repeated updates and keeps output-token cost low even as the estimated cost without reuse increases. Across progress buckets, the estimated output-token cost without reuse is $8.1$--$32.6\times$ the actual cost, highlighting the savings from keeping reusable memories synchronized throughout a long narrative. To examine how effectively each method uses additional capacity, we vary the bookmark limit in \our over $\{2,3,5\}$ and the maximum profile-token capacity specified in ETA's prompt over $\{128,256,512,1024\}$ (3-way for $\{128,256,512\}$ each section). Figure~\ref{fig:grounding_efficiency} (right) shows that allowing more bookmarks consistently improves predictions, whereas increasing (3-way) ETA's profile capacity yields diminishing gains and eventually a slight decline. This pattern supports organizing grounding around multiple task-relevant memories as an effective way to obtain both better predictions at lower cost and further gains when more resources are available.

\subsection{Ablation Study}

\begin{figure}[htbp]
    \centering
    \includegraphics[width=\linewidth]{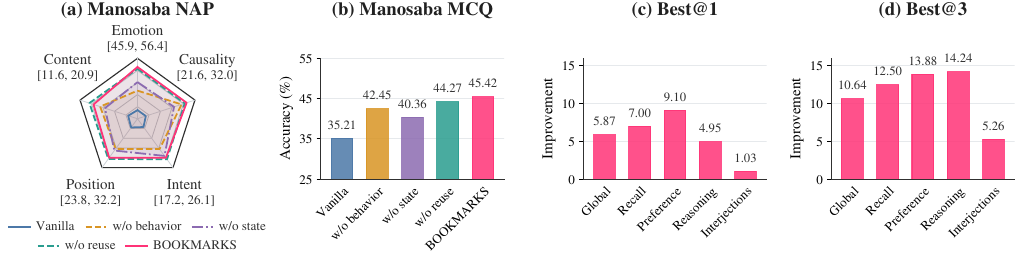}
    \caption{Manosaba ablations on (a) Best@1 NAP, with axes linearly rescaled to the labeled ranges, and (b) MCQ accuracy. (c,d) NAP improvements of \our over Vanilla (percentage points), overall and by action type, under Best@1 and Best@3, respectively.}
    \label{fig:ablation_action_types}
\end{figure}

The Manosaba ablations (1) evaluate the contributions of state and behavioral bookmarks to grounding and (2) test whether reuse accelerates memory updates without compromising prediction quality. As shown in Figure~\ref{fig:ablation_action_types}(a,b), removing either memory type lowers performance across all five NAP dimensions and on MCQ. The NAP losses reflect their different roles, as removing behavioral bookmarks hurts \textit{Emotion} and \textit{Intent} more, whereas removing state bookmarks has a larger effect on \textit{Causality} and \textit{Content}. This pattern is consistent with behavioral examples grounding how a character responds and state traces preserving the evolving circumstances behind an action. On MCQ, removing state and behavioral bookmarks reduces accuracy by $5.06$ and $2.97$ percentage points, respectively, further supporting the value of both memory types for reasoning over storyline evidence. Disabling reuse yields performance comparable to the full framework on both NAP and MCQ. Combined with the cost savings in Figure~\ref{fig:grounding_efficiency}, this supports reuse as an efficiency mechanism that preserves grounding quality, while the specialized memories supply the complementary evidence needed for stronger predictions.

\subsection{Improvement Analysis}

Figure~\ref{fig:ablation_action_types}(c,d) further divides improvement gain by action type, which shows that \our's gains are concentrated in actions that depend on storyline knowledge. \textit{Recall} and \textit{Preference} both improve more than the global average under Best@1 and Best@3, highlighting the contribution of remembered events and character-specific preferences. Trivial \textit{Interjections} benefit much less, as they typically require little storyline evidence. The substantially larger gains on \textit{Reasoning} under Best@3 suggest that Best@1 underestimates \our's contribution as a memory system, since performance also depends on the model's ability to derive the final conclusion. Best@3 better reveals the value of the reasoning-relevant information \our collects, supporting targeted access to storyline evidence as the main source of its gains.

\section{Conclusion and Future Work}

We present \our, a search-based memory framework that improves role-playing fidelity through active grounding and accelerates this process through passive updating. We show that \our outperforms retrieval-based grounding and incremental profiling on next-action prediction and storyline reasoning. As future work, we plan to extend \our to character-specific knowledge tracking, integrate it with self-refinement, and develop update policies tailored to different bookmark types and narrative structures.

\section*{AI use statement}
\label{apdx:ai_use_statement}
We used large language models (LLMs) only for copy-editing: improving grammar, clarity, and overall writing style to enhance readability. The LLMs did not contribute to or modify the scientific content, core ideas, methodology, or experimental results. The authors take full responsibility for the manuscript’s final content and accuracy.

\bibliography{custom}

@inproceedings{codification,
  title     = {Codifying Character Logic in Role-Playing},
  author    = {Letian Peng and Jingbo Shang},
  booktitle = {Proceedings of the Thirty-Ninth Annual Conference on Neural Information Processing Systems (NeurIPS 2025) Poster Session},
  year      = {2025},
  location  = {San Diego, CA, USA},
  note      = {Poster presentation, NeurIPS 2025},
  url       = {https://neurips.cc/virtual/2025/loc/san-diego/poster/117989}
}

@inproceedings{search-r1,
  title={Search-R1: Training LLMs to Reason and Leverage Search Engines with Reinforcement Learning},
  author={Jin, Bowen and Zeng, Hansi and Yue, Zhenrui and Yoon, Jinsung and Arik, Sercan O and Wang, Dong and Zamani, Hamed and Han, Jiawei},
  booktitle={Second Conference on Language Modeling},
  year={2025}
}

@inproceedings{ricl,
    title = "Learning to Retrieve In-Context Examples for Large Language Models",
    author = "Wang, Liang  and
      Yang, Nan  and
      Wei, Furu",
    editor = "Graham, Yvette  and
      Purver, Matthew",
    booktitle = "Proceedings of the 18th Conference of the European Chapter of the Association for Computational Linguistics (Volume 1: Long Papers)",
    month = mar,
    year = "2024",
    address = "St. Julian{'}s, Malta",
    publisher = "Association for Computational Linguistics",
    url = "https://aclanthology.org/2024.eacl-long.105/",
    doi = "10.18653/v1/2024.eacl-long.105",
    pages = "1752--1767"
}

@inproceedings{SweetieChat,
  author       = {Jing Ye and
                  Lu Xiang and
                  Yaping Zhang and
                  Chengqing Zong},
  editor       = {Owen Rambow and
                  Leo Wanner and
                  Marianna Apidianaki and
                  Hend Al{-}Khalifa and
                  Barbara Di Eugenio and
                  Steven Schockaert},
  title        = {SweetieChat: {A} Strategy-Enhanced Role-playing Framework for Diverse
                  Scenarios Handling Emotional Support Agent},
  booktitle    = {Proceedings of the 31st International Conference on Computational
                  Linguistics, {COLING} 2025, Abu Dhabi, UAE, January 19-24, 2025},
  pages        = {4646--4669},
  publisher    = {Association for Computational Linguistics},
  year         = {2025},
  url          = {https://aclanthology.org/2025.coling-main.312/},
  bibsource    = {dblp computer science bibliography, https://dblp.org}
}

@article{long_form_story_reasoning,
  author       = {Alexander Gurung and
                  Mirella Lapata},
  title        = {Learning to Reason for Long-Form Story Generation},
  journal      = {CoRR},
  volume       = {abs/2503.22828},
  year         = {2025},
  url          = {https://doi.org/10.48550/arXiv.2503.22828},
  doi          = {10.48550/ARXIV.2503.22828},
  eprinttype    = {arXiv},
  eprint       = {2503.22828},
  bibsource    = {dblp computer science bibliography, https://dblp.org}
}

@article{RPGBenchmark,
  author       = {Pengfei Yu and
                  Dongming Shen and
                  Silin Meng and
                  Jaewon Lee and
                  Weisu Yin and
                  Andrea Yaoyun Cui and
                  Zhenlin Xu and
                  Yi Zhu and
                  Xingjian Shi and
                  Mu Li and
                  Alex Smola},
  title        = {{RPGBENCH:} Evaluating Large Language Models as Role-Playing Game
                  Engines},
  journal      = {CoRR},
  volume       = {abs/2502.00595},
  year         = {2025},
  url          = {https://doi.org/10.48550/arXiv.2502.00595},
  doi          = {10.48550/ARXIV.2502.00595},
  eprinttype    = {arXiv},
  eprint       = {2502.00595},
  bibsource    = {dblp computer science bibliography, https://dblp.org}
}

@inproceedings{emotion,
    title = "{CARER}: Contextualized Affect Representations for Emotion Recognition",
    author = "Saravia, Elvis  and
      Liu, Hsien-Chi Toby  and
      Huang, Yen-Hao  and
      Wu, Junlin  and
      Chen, Yi-Shin",
    booktitle = "Proceedings of the 2018 Conference on Empirical Methods in Natural Language Processing",
    month = oct # "-" # nov,
    year = "2018",
    address = "Brussels, Belgium",
    publisher = "Association for Computational Linguistics",
    url = "https://www.aclweb.org/anthology/D18-1404",
    doi = "10.18653/v1/D18-1404",
    pages = "3687--3697",
}

@inproceedings{shao2023character,
  title={Character-LLM: A Trainable Agent for Role-Playing},
  author={Shao, Yunfan and Li, Linyang and Dai, Junqi and Qiu, Xipeng},
  booktitle={Proceedings of the 2023 Conference on Empirical Methods in Natural Language Processing},
  pages={13153--13187},
  year={2023}
}

@article{chenpersona,
  title={From Persona to Personalization: A Survey on Role-Playing Language Agents},
  author={Chen, Jiangjie and Wang, Xintao and Xu, Rui and Yuan, Siyu and Zhang, Yikai and Shi, Wei and Xie, Jian and Li, Shuang and Yang, Ruihan and Zhu, Tinghui and others},
  journal={Transactions on Machine Learning Research},
  year={2024}
}

@article{yan2023larp,
  title={Larp: Language-agent role play for open-world games},
  author={Yan, Ming and Li, Ruihao and Zhang, Hao and Wang, Hao and Yang, Zhilan and Yan, Ji},
  journal={arXiv preprint arXiv:2312.17653},
  year={2023}
}

@article{moore2024rolellm,
  title={Rolellm: benchmarking, eliciting, and enhancing role-playing abilities of large language models},
  author={Moore Wang, Zekun and Peng, Zhongyuan and Que, Haoran and Liu, Jiaheng and Zhou, Wangchunshu and Wu, Yuhan and Guo, Hongcheng and Gan, Ruitong and Ni, Zehao and Yang, Jian and others},
  journal={Findings of the Association for Computational Linguistics: ACL 2024},
  pages={14743--14777},
  year={2024},
  publisher={Association for Computational Linguistics}
}

@inproceedings{yu2024neeko,
  title={Neeko: Leveraging Dynamic LoRA for Efficient Multi-Character Role-Playing Agent},
  author={Yu, Xiaoyan and Luo, Tongxu and Wei, Yifan and Lei, Fangyu and Huang, Yiming and Peng, Hao and Zhu, Liehuang},
  booktitle={Proceedings of the 2024 Conference on Empirical Methods in Natural Language Processing},
  pages={12540--12557},
  year={2024}
}

@article{tang2025thinking,
  title={Thinking in Character: Advancing Role-Playing Agents with Role-Aware Reasoning},
  author={Tang, Yihong and Chen, Kehai and Yang, Muyun and Niu, Zhengyu and Li, Jing and Zhao, Tiejun and Zhang, Min},
  journal={arXiv preprint arXiv:2506.01748},
  year={2025}
}

@article{cheng2025psymem,
  title={PsyMem: Fine-grained psychological alignment and Explicit Memory Control for Advanced Role-Playing LLMs},
  author={Cheng, Xilong and Qin, Yunxiao and Tan, Yuting and Li, Zhengnan and Wang, Ye and Xiao, Hongjiang and Zhang, Yuan},
  journal={arXiv preprint arXiv:2505.12814},
  year={2025}
}

@inproceedings{li2024graphreader,
  title={GraphReader: Building Graph-based Agent to Enhance Long-Context Abilities of Large Language Models},
  author={Li, Shilong and He, Yancheng and Guo, Hangyu and Bu, Xingyuan and Bai, Ge and Liu, Jie and Liu, Jiaheng and Qu, Xingwei and Li, Yangguang and Ouyang, Wanli and others},
  booktitle={Findings of the Association for Computational Linguistics: EMNLP 2024},
  pages={12758--12786},
  year={2024}
}

@article{sun2024identity,
  title={Identity-driven hierarchical role-playing agents},
  author={Sun, Libo and Wang, Siyuan and Huang, Xuanjing and Wei, Zhongyu},
  journal={arXiv preprint arXiv:2407.19412},
  year={2024}
}

@inproceedings{yoneda2024statler,
  title={Statler: State-maintaining language models for embodied reasoning},
  author={Yoneda, Takuma and Fang, Jiading and Li, Peng and Zhang, Huanyu and Jiang, Tianchong and Lin, Shengjie and Picker, Ben and Yunis, David and Mei, Hongyuan and Walter, Matthew R},
  booktitle={2024 IEEE International Conference on Robotics and Automation (ICRA)},
  pages={15083--15091},
  year={2024},
  organization={IEEE}
}

@inproceedings{bookworld,
  author       = {Yiting Ran and
                  Xintao Wang and
                  Tian Qiu and
                  Jiaqing Liang and
                  Yanghua Xiao and
                  Deqing Yang},
  editor       = {Wanxiang Che and
                  Joyce Nabende and
                  Ekaterina Shutova and
                  Mohammad Taher Pilehvar},
  title        = {{BOOKWORLD:} From Novels to Interactive Agent Societies for Story
                  Creation},
  booktitle    = {Proceedings of the 63rd Annual Meeting of the Association for Computational
                  Linguistics (Volume 1: Long Papers), {ACL} 2025, Vienna, Austria,
                  July 27 - August 1, 2025},
  pages        = {15898--15912},
  publisher    = {Association for Computational Linguistics},
  year         = {2025},
  url          = {https://aclanthology.org/2025.acl-long.773/},
  bibsource    = {dblp computer science bibliography, https://dblp.org}
}

@inproceedings{in_character,
  author       = {Xintao Wang and
                  Yunze Xiao and
                  Jen{-}tse Huang and
                  Siyu Yuan and
                  Rui Xu and
                  Haoran Guo and
                  Quan Tu and
                  Yaying Fei and
                  Ziang Leng and
                  Wei Wang and
                  Jiangjie Chen and
                  Cheng Li and
                  Yanghua Xiao},
  editor       = {Lun{-}Wei Ku and
                  Andre Martins and
                  Vivek Srikumar},
  title        = {InCharacter: Evaluating Personality Fidelity in Role-Playing Agents
                  through Psychological Interviews},
  booktitle    = {Proceedings of the 62nd Annual Meeting of the Association for Computational
                  Linguistics (Volume 1: Long Papers), {ACL} 2024, Bangkok, Thailand,
                  August 11-16, 2024},
  pages        = {1840--1873},
  publisher    = {Association for Computational Linguistics},
  year         = {2024},
  url          = {https://doi.org/10.18653/v1/2024.acl-long.102},
  doi          = {10.18653/V1/2024.ACL-LONG.102},
  bibsource    = {dblp computer science bibliography, https://dblp.org}
}

@article{coser,
  author       = {Xintao Wang and
                  Heng Wang and
                  Yifei Zhang and
                  Xinfeng Yuan and
                  Rui Xu and
                  Jen{-}tse Huang and
                  Siyu Yuan and
                  Haoran Guo and
                  Jiangjie Chen and
                  Wei Wang and
                  Yanghua Xiao and
                  Shuchang Zhou},
  title        = {CoSER: Coordinating LLM-Based Persona Simulation of Established Roles},
  journal      = {CoRR},
  volume       = {abs/2502.09082},
  year         = {2025},
  url          = {https://doi.org/10.48550/arXiv.2502.09082},
  doi          = {10.48550/ARXIV.2502.09082},
  eprinttype    = {arXiv},
  eprint       = {2502.09082},
  bibsource    = {dblp computer science bibliography, https://dblp.org}
}

@misc{survey_rp,
      title={From Persona to Personalization: A Survey on Role-Playing Language Agents}, 
      author={Jiangjie Chen and Xintao Wang and Rui Xu and Siyu Yuan and Yikai Zhang and Wei Shi and Jian Xie and Shuang Li and Ruihan Yang and Tinghui Zhu and Aili Chen and Nianqi Li and Lida Chen and Caiyu Hu and Siye Wu and Scott Ren and Ziquan Fu and Yanghua Xiao},
      year={2024},
      eprint={2404.18231},
      archivePrefix={arXiv},
      primaryClass={cs.CL},
      url={https://arxiv.org/abs/2404.18231}, 
}

@article{liu2024grounded,
  title={Grounded answers for multi-agent decision-making problem through generative world model},
  author={Liu, Zeyang and Yang, Xinrui and Sun, Shiguang and Qian, Long and Wan, Lipeng and Chen, Xingyu and Lan, Xuguang},
  journal={Advances in Neural Information Processing Systems},
  volume={37},
  pages={46622--46652},
  year={2024}
}

@inproceedings{tseng2024two,
  title={Two tales of persona in llms: A survey of role-playing and personalization},
  author={Tseng, Yu-Min and Huang, Yu-Chao and Hsiao, Teng-Yun and Chen, Wei-Lin and Huang, Chao-Wei and Meng, Yu and Chen, Yun-Nung},
  booktitle={Findings of the Association for Computational Linguistics: EMNLP 2024},
  pages={16612--16631},
  year={2024}
}

@article{peng2026deriving,
  title={Deriving Character Logic from Storyline as Codified Decision Trees},
  author={Peng, Letian and Zhou, Kun and Yun, Longfei and Hou, Yupeng and Shang, Jingbo},
  journal={arXiv preprint arXiv:2601.10080},
  year={2026}
}

@article{wang2026memory,
  title={Memory-Driven Role-Playing: Evaluation and Enhancement of Persona Knowledge Utilization in LLMs},
  author={Wang, Kai and You, Haoyang and Zhang, Yang and Wang, Zhongjie},
  journal={arXiv preprint arXiv:2603.19313},
  year={2026}
}

@article{yang2025psyplay,
  title={Psyplay: Personality-infused role-playing conversational agents},
  author={Yang, Tao and Zhu, Yuhua and Quan, Xiaojun and Liu, Cong and Wang, Qifan},
  journal={arXiv preprint arXiv:2502.03821},
  year={2025}
}

@inproceedings{zhoupersonaeval,
  title={PersonaEval: Are LLM Evaluators Human Enough to Judge Role-Play?},
  author={Zhou, Lingfeng and Zhang, Jialing and Gao, Jin and Jiang, Mohan and Wang, Dequan},
  booktitle={Second Conference on Language Modeling},
  year={2025}
}

@inproceedings{zhousotopia,
  title={SOTOPIA: Interactive Evaluation for Social Intelligence in Language Agents},
  author={Zhou, Xuhui and Zhu, Hao and Mathur, Leena and Zhang, Ruohong and Yu, Haofei and Qi, Zhengyang and Morency, Louis-Philippe and Bisk, Yonatan and Fried, Daniel and Neubig, Graham and others},
  booktitle={The Twelfth International Conference on Learning Representations},
  year={2024}
}

@inproceedings{chen2024socialbench,
  title={Socialbench: Sociality evaluation of role-playing conversational agents},
  author={Chen, Hongzhan and Chen, Hehong and Yan, Ming and Xu, Wenshen and Xing, Gao and Shen, Weizhou and Quan, Xiaojun and Li, Chenliang and Zhang, Ji and Huang, Fei},
  booktitle={Findings of the Association for Computational Linguistics: ACL 2024},
  pages={2108--2126},
  year={2024}
}

@inproceedings{ahn2024timechara,
  title={Timechara: Evaluating point-in-time character hallucination of role-playing large language models},
  author={Ahn, Jaewoo and Lee, Taehyun and Lim, Junyoung and Kim, Jin-Hwa and Yun, Sangdoo and Lee, Hwaran and Kim, Gunhee},
  booktitle={Findings of the Association for Computational Linguistics: ACL 2024},
  pages={3291--3325},
  year={2024}
}

@inproceedings{sadeq2024mitigating,
  title={Mitigating hallucination in fictional character role-play},
  author={Sadeq, Nafis and Xie, Zhouhang and Kang, Byungkyu and Lamba, Prarit and Gao, Xiang and McAuley, Julian},
  booktitle={Findings of the Association for Computational Linguistics: EMNLP 2024},
  pages={14467--14479},
  year={2024}
}

@article{shen2023roleeval,
  title={Roleeval: A bilingual role evaluation benchmark for large language models},
  author={Shen, Tianhao and Li, Sun and Tu, Quan and Xiong, Deyi},
  journal={arXiv preprint arXiv:2312.16132},
  year={2023}
}

@inproceedings{tu2024charactereval,
  title={Charactereval: A chinese benchmark for role-playing conversational agent evaluation},
  author={Tu, Quan and Fan, Shilong and Tian, Zihang and Shen, Tianhao and Shang, Shuo and Gao, Xin and Yan, Rui},
  booktitle={Proceedings of the 62nd Annual Meeting of the Association for Computational Linguistics (Volume 1: Long Papers)},
  pages={11836--11850},
  year={2024}
}

@inproceedings{lu2025rolemrc,
  title={Rolemrc: A fine-grained composite benchmark for role-playing and instruction-following},
  author={Lu, Junru and Li, Jiazheng and Shen, Guodong and Gui, Lin and An, Siyu and He, Yulan and Yin, Di and Sun, Xing},
  booktitle={Findings of the Association for Computational Linguistics: ACL 2025},
  pages={21008--21030},
  year={2025}
}

@inproceedings{he2025crab,
  title={Crab: A novel configurable role-playing llm with assessing benchmark},
  author={He, Kai and Huang, Yucheng and Wang, Wenqing and Ran, Delong and Sheng, Dongming and Huang, Junxuan and Lin, Qika and Xu, Jiaxing and Liu, Wenqiang and Feng, Mengling},
  booktitle={Proceedings of the 63rd Annual Meeting of the Association for Computational Linguistics (Volume 1: Long Papers)},
  pages={15030--15052},
  year={2025}
}

@article{ding2025rolermbench,
  title={Rolermbench \& rolerm: Towards reward modeling for profile-based role play in dialogue systems},
  author={Ding, Hang and Feng, Qiming and Liu, Dongqi and Zhao, Qi and Yao, Tao and Wang, Shuo and Chen, Dongsheng and Li, Jian and Gan, Zhenye and Zhang, Jiangning and others},
  journal={arXiv preprint arXiv:2512.10575},
  year={2025}
}

@inproceedings{lu2024large,
  title={Large language models are superpositions of all characters: Attaining arbitrary role-play via self-alignment},
  author={Lu, Keming and Yu, Bowen and Zhou, Chang and Zhou, Jingren},
  booktitle={Proceedings of the 62nd Annual Meeting of the Association for Computational Linguistics (Volume 1: Long Papers)},
  pages={7828--7840},
  year={2024}
}

@article{wang2025opencharacter,
  title={Opencharacter: Training customizable role-playing llms with large-scale synthetic personas},
  author={Wang, Xiaoyang and Zhang, Hongming and Ge, Tao and Yu, Wenhao and Yu, Dian and Yu, Dong},
  journal={arXiv preprint arXiv:2501.15427},
  year={2025}
}

@inproceedings{yang-etal-2025-crafting,
    title = "Crafting Customisable Characters with {LLM}s: A Persona-Driven Role-Playing Agent Framework",
    author = "Yang, Bohao  and
      Liu, Dong  and
      Xiao, Chenghao  and
      Zhao, Kun  and
      Tang, Chen  and
      Li, Chao  and
      Yuan, Lin  and
      Guang, Yang  and
      Lin, Chenghua",
    editor = "Christodoulopoulos, Christos  and
      Chakraborty, Tanmoy  and
      Rose, Carolyn  and
      Peng, Violet",
    booktitle = "Findings of the Association for Computational Linguistics: EMNLP 2025",
    month = nov,
    year = "2025",
    address = "Suzhou, China",
    publisher = "Association for Computational Linguistics",
    url = "https://aclanthology.org/2025.findings-emnlp.1100/",
    doi = "10.18653/v1/2025.findings-emnlp.1100",
    pages = "20216--20240",
    ISBN = "979-8-89176-335-7"
}

@article{tang2024erabal,
  title={Erabal: Enhancing role-playing agents through boundary-aware learning},
  author={Tang, Yihong and Ou, Jiao and Liu, Che and Zhang, Fuzheng and Zhang, Di and Gai, Kun},
  journal={arXiv preprint arXiv:2409.14710},
  year={2024}
}

@article{wang2025raiden,
  title={Raiden-r1: Improving role-awareness of llms via grpo with verifiable reward},
  author={Wang, Zongsheng and Sun, Kaili and Wu, Bowen and Yu, Qun and Li, Ying and Wang, Baoxun},
  journal={arXiv preprint arXiv:2505.10218},
  year={2025}
}

@article{fang2025charm,
  title={Charm: Character-based act-adaptive reward modeling for advanced role-playing language agents},
  author={Fang, Feiteng and Lin, Ting-En and Wu, Yuchuan and Liu, Xiong and Huang, Xiang and Chen, Dingwei and Ye, Jing and Zhang, Haonan and Zhu, Liang and Alinejad-Rokny, Hamid and others},
  journal={arXiv e-prints},
  pages={arXiv--2505},
  year={2025}
}

@inproceedings{liu2025cogdual,
  title={CogDual: Enhancing Dual Cognition of LLMs via Reinforcement Learning with Implicit Rule-Based Rewards},
  author={Liu, Cheng and Lu, Yifei and Ye, Fanghua and Li, Jian and Chen, Xingyu and Ren, Feiliang and Tu, Zhaopeng and Li, Xiaolong},
  booktitle={Proceedings of the 2025 Conference on Empirical Methods in Natural Language Processing},
  pages={27295--27324},
  year={2025}
}

@inproceedings{ji2025enhancing,
  title={Enhancing persona consistency for llms’ role-playing using persona-aware contrastive learning},
  author={Ji, Ke and Lian, Yixin and Li, Linxu and Gao, Jingsheng and Li, Weiyuan and Dai, Bin},
  booktitle={Findings of the Association for Computational Linguistics: ACL 2025},
  pages={26221--26238},
  year={2025}
}

@article{chen2025persona,
  title={Persona vectors: Monitoring and controlling character traits in language models},
  author={Chen, Runjin and Arditi, Andy and Sleight, Henry and Evans, Owain and Lindsey, Jack},
  journal={arXiv preprint arXiv:2507.21509},
  year={2025}
}

@article{wang2025recursively,
  title={Recursively summarizing enables long-term dialogue memory in large language models},
  author={Wang, Qingyue and Fu, Yanhe and Cao, Yanan and Wang, Shuai and Tian, Zhiliang and Ding, Liang},
  journal={Neurocomputing},
  volume={639},
  pages={130193},
  year={2025},
  publisher={Elsevier}
}

@article{lu2023memochat,
  title={Memochat: Tuning llms to use memos for consistent long-range open-domain conversation},
  author={Lu, Junru and An, Siyu and Lin, Mingbao and Pergola, Gabriele and He, Yulan and Yin, Di and Sun, Xing and Wu, Yunsheng},
  journal={arXiv preprint arXiv:2308.08239},
  year={2023}
}

@inproceedings{zhong2024memorybank,
  title={Memorybank: Enhancing large language models with long-term memory},
  author={Zhong, Wanjun and Guo, Lianghong and Gao, Qiqi and Ye, He and Wang, Yanlin},
  booktitle={Proceedings of the AAAI conference on artificial intelligence},
  volume={38},
  pages={19724--19731},
  year={2024}
}

@article{wang2025rolerag,
  title={Rolerag: Enhancing llm role-playing via graph guided retrieval},
  author={Wang, Yongjie and Leung, Jonathan and Shen, Zhiqi},
  journal={arXiv preprint arXiv:2505.18541},
  year={2025}
}

@article{chen2025moom,
  title={Moom: Maintenance, organization and optimization of memory in ultra-long role-playing dialogues},
  author={Chen, Weishu and Tang, Jinyi and Hou, Zhouhui and Han, Shihao and Zhan, Mingjie and Huang, Zhiyuan and Liu, Delong and Guo, Jiawei and Zhao, Zhicheng and Su, Fei},
  journal={arXiv preprint arXiv:2509.11860},
  year={2025}
}

@inproceedings{huang2024emotional,
  title={Emotional RAG: Enhancing role-playing agents through emotional retrieval},
  author={Huang, Le and Lan, Hengzhi and Sun, Zijun and Shi, Chuan and Bai, Ting},
  booktitle={2024 IEEE International Conference on Knowledge Graph (ICKG)},
  pages={120--127},
  year={2024},
  organization={IEEE}
}

@article{yi2025score,
  title={Score: Story coherence and retrieval enhancement for ai narratives},
  author={Yi, Qiang and He, Yangfan and Wang, Jianhui and Song, Xinyuan and Qian, Shiyao and Yuan, Xinhang and Xin, Yi and Wang, Yijin and Tang, Jingqun and Li, Yuchen and others},
  journal={arXiv preprint arXiv:2503.23512},
  year={2025}
}

@inproceedings{xia2025storywriter,
  title={Storywriter: A multi-agent framework for long story generation},
  author={Xia, Haotian and Peng, Hao and Qi, Yunjia and Xu, Bin and Li, Juanzi and Lei, Hou and Wang, Xiaozhi},
  booktitle={Proceedings of the 34th ACM International Conference on Information and Knowledge Management},
  pages={6559--6563},
  year={2025}
}
\bibliographystyle{iclr2027_conference}

\clearpage

\appendix

\section{Limitations}
\label{apdx:limitations}

This work focuses on establishing \our as a general search-based memory framework for long-horizon role-playing, while leaving several extensions for future exploration. First, the current framework mainly maintains storyline-level states, behaviors, and concepts, and can be further extended to finer-grained recognition management, such as explicitly tracking which character knows which information at each story point. Second, \our is currently used as a grounding module before generation, while future work may integrate it with self-refinement so that agents can revise generated actions when bookmark evidence reveals inconsistencies. Finally, although we design type-specific update procedures for concept, state, and behavioral bookmarks, more customized update policies could be developed for different query types, characters, and narrative structures.



\section{Prompts}
\label{apdx:prompts}

\providecommand{\promptpositiongloss}{\raisebox{-1.3pt}{\includegraphics[height=10pt]{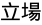}}}

Tables~\ref{tab:bookmark_prompts}--\ref{tab:evaluation_prompts} present the prompts for bookmark construction and updating, next-action prediction, and evaluation. Braced fields denote runtime inputs. Each evaluation dimension receives an independent yes/no judgment, scored as $100/0$.

\begin{table}[htbp]
\caption{Bookmark proposal, matching, and concept-update prompts.}
\label{tab:bookmark_prompts}
\centering
\footnotesize
\setlength{\tabcolsep}{0pt}
\begin{tabular}{@{}>{\raggedright\arraybackslash}p{\linewidth}@{}}
\toprule
\textbf{Section proposal} \\[2pt]
\# Scene\newline{}\{scene\}
\par\vspace{2pt}
It is now \{target\_character\}'s turn to make the next utterance. Before deciding it, propose up to \{max\_k\} bookmarks that may bring useful information from the preceding story into the current context.
\par\vspace{2pt}
A concept bookmark names a concrete concept or entity and searches that name directly. A state bookmark keeps a transition trace of natural-language state nodes for a character, relationship, object, place, plan, or shared environment.
\par\vspace{2pt}
\# Bookmarks already visible in the scene\newline{}\{visible\_catalog\}
\par\vspace{2pt}
\# Other bookmark history\newline{}\{nonvisible\_history\}
\par\vspace{2pt}
\# Characters that may own a bookmark\newline{}\{character\_options\}
\par\vspace{2pt}
Use short noun-phrase section names. Write every "section" value in the same language as the input scene. A bookmark may concern another character when that helps with \{target\_character\}'s continuation; use "env" only for shared or environmental state. Avoid repeating information already supplied by the visible scene or its bookmarks. Return fewer than \{max\_k\}, or an empty list, when there are fewer useful searches.
\par\vspace{2pt}
Return only a JSON array. Each item must contain "section", "tag", and "character", and the tag must be "concept", "state", or "behavioral".\newline{}Concept example: \{"section": "Project Polaris", "tag": "concept", "character": "\{target\_character\}"\}\newline{}State example: \{"section": "Band members' current location", "tag": "state", "character": "env"\}\newline{}Behavioral example: \{"section": "\{target\_character\}'s response pattern when being forcefully invited", "tag": "behavioral", "character": "\{target\_character\}"\} \\[3pt]
\midrule
\textbf{Bookmark matching} \\[2pt]
You are choosing whether one existing bookmark can be reused for a new bookmark.
\par\vspace{2pt}
New bookmark section:\newline{}\{new\_section\}
\par\vspace{2pt}
Candidate existing bookmark sections:\newline{}\{candidate\_catalog\}
\par\vspace{2pt}
Choose one candidate only when its section is semantically equivalent to the new section and both sections maintain the same bookmark slot over time. A section that is merely related, partially overlapping, broader, narrower, or otherwise useful must not be reused.
\par\vspace{2pt}
Output only the exact bookmark number of the single best reuse candidate, or none if no candidate has the same section meaning. \\[3pt]
\midrule
\textbf{Concept update} \\[2pt]
The bookmark section is "\{section\}". Here are the places where that name appears:\newline{}\{mention\_context\}
\par\vspace{2pt}
The current answer is:\newline{}\{answer\}
\par\vspace{2pt}
Write a short, self-contained memory note that names "\{section\}" and explains it using the information above, keeping useful earlier details when they still apply. Return only the note. \\[3pt]
\bottomrule
\end{tabular}
\end{table}

\begin{table}[htbp]
\caption{State-transition check and state-node writing prompts.}
\label{tab:state_prompts}
\centering
\footnotesize
\setlength{\tabcolsep}{0pt}
\begin{tabular}{@{}>{\raggedright\arraybackslash}p{\linewidth}@{}}
\toprule
\textbf{State-transition check} \\[2pt]
The state bookmark "\{section\}" belongs to \{state\_character\}.
\par\vspace{0.5pt}
\# Recent state-transition trace\newline{}\{recent\_trace\}
\par\vspace{0.5pt}
\{empty\_trace\_note\}
\par\vspace{0.5pt}
\{background\_note\}\newline{}\# Current story chunk --- the only transition evidence\newline{}\{chunk\}
\par\vspace{0.5pt}
\{direct\_action\_note\}
\par\vspace{0.5pt}
Each state node is one natural-language state note. Its applicability conditions---such as supported time, place, event phase, relationship phase, plan phase, or other situational context---are written directly inside that same note. There is no separate precondition field.
\par\vspace{0.5pt}
Choose exactly one operation:
\par\vspace{0.5pt}
none\newline{}- The current chunk gives no new relevant evidence, merely repeats the current state, or only the context-only background is relevant.
\par\vspace{0.5pt}
modify\_current\newline{}- The current chunk refines, corrects, narrows, or adds information to the latest state under the same applicability conditions already written in that state.\newline{}- Rewrite the current node rather than adding a transition.\newline{}- This operation is invalid when the trace is empty.
\par\vspace{0.5pt}
append\_state\newline{}- The current chunk establishes a successor state whose meaningful applicability conditions differ from the current node---for example a new time, place, event phase, relationship phase, plan phase, or situation.\newline{}- Use this for the first established state when the trace is empty.\newline{}- Do not append merely because story time passed or the location changed; the chunk must establish or change this bookmark's state under those new conditions.
\par\vspace{0.5pt}
For a character-owned bookmark, choose none unless the current chunk contains information explicitly about that character or unambiguously referring to them; never transfer another character's state. Here "env" means shared or environmental state.
\par\vspace{0.5pt}
Return exactly one token: none, modify\_current, or append\_state. \\[3pt]
\midrule
\textbf{State-node writing} \\[2pt]
The state bookmark "\{section\}" belongs to \{state\_character\}.
\par\vspace{0.5pt}
\# Recent state-transition trace\newline{}\{recent\_trace\}
\par\vspace{0.5pt}
\# Applied operation\newline{}\{operation\}
\par\vspace{0.5pt}
\{background\_note\}\newline{}\# Current story chunk --- the only transition evidence\newline{}\{chunk\}
\par\vspace{0.5pt}
\{direct\_action\_note\}
\par\vspace{0.5pt}
\{operation\_instruction\}
\par\vspace{0.5pt}
The returned state must be short and self-contained. Write its applicability conditions directly inside the state sentence itself: naturally state when, where, and under what event, interaction, investigation, relationship, plan, or other situational phase it applies whenever those conditions are supported. Conditions must be part of the state text, not separate fields, labels, JSON, or metadata. Omit unsupported time or place rather than inventing them, but always make the relevant situation or phase clear enough for a later model to decide whether to modify this node or append a successor.
\par\vspace{0.5pt}
Use the context-only background only to resolve references. Do not copy facts from it as new state evidence. Here "env" means shared or environmental state.
\par\vspace{0.5pt}
Return only the natural-language state note. \\[3pt]
\midrule
\textbf{Instruction for modify\_current} \\[2pt]
Rewrite the latest state node as one complete natural-language state note. Preserve earlier information that remains valid, and keep it within the same applicability conditions. Do not move it into a distinct time, place, phase, or situation; that would require append\_state. \\[3pt]
\midrule
\textbf{Instruction for append\_state} \\[2pt]
Write one new natural-language state note that succeeds the existing trace. Do not rewrite an earlier node. If the previous state ended but no replacement is known, explicitly state that under the new conditions instead of inventing a value. \\[3pt]
\bottomrule
\end{tabular}
\end{table}

\begin{table}[htbp]
\caption{Behavioral annotation and filtering prompts.}
\label{tab:behavioral_prompts}
\centering
\footnotesize
\setlength{\tabcolsep}{0pt}
\begin{tabular}{@{}>{\raggedright\arraybackslash}p{\linewidth}@{}}
\toprule
\textbf{Scene--action annotation} \\[2pt]
\# Scene\newline{}\{scene\}
\par\vspace{2pt}
\# Action\newline{}\{action\}
\par\vspace{2pt}
In one short sentence, explain what in the preceding scene caused or prompted this action. Use any concept or state bookmark annotations only when needed to make that causal link clear. This explanation is independent of every behavioral bookmark section and target, so do not judge whether the action belongs to any behavioral pattern.
\par\vspace{2pt}
Return only that sentence in the same language as the action. \\[3pt]
\midrule
\textbf{Behavioral filtering} \\[2pt]
The behavioral bookmark "\{section\}" collects \{character\}'s actions that fit that section.
\par\vspace{2pt}
\{candidate\_text\}
\par\vspace{2pt}
Each action is followed by a short parenthetical \textreferencemark{} annotation explaining what in the preceding scene caused or prompted it. The annotation is independent of this behavioral bookmark. For each candidate, answer "yes" if the annotated action belongs in this bookmark and "no" otherwise. This step is only collecting matching actions; one action does not need to prove the overall pattern by itself.
\par\vspace{2pt}
Return only a JSON list of \{n\_candidates\} "yes" or "no" strings in candidate order. \\[3pt]
\bottomrule
\end{tabular}
\end{table}

\begin{table}[htbp]
\caption{Next-action prediction prompts, shown for spoken utterances.}
\label{tab:prediction_prompts}
\centering
\footnotesize
\setlength{\tabcolsep}{0pt}
\begin{tabular}{@{}>{\raggedright\arraybackslash}p{\linewidth}@{}}
\toprule
\textbf{Single-action prediction} \\[2pt]
\# Background Summary\newline{}\{grounding\}\newline{}\texttt{\char45\char45\char45}\newline{}\# Current Scene\newline{}\{scene\}\newline{}\texttt{\char45\char45\char45}\newline{}Predict the next utterance spoken by \{character\} in the following scene. The "action" field must contain a concise spoken line by \{character\}, formatted as "\{character\}: ...", not narration, a nonverbal action, or another character's line.
\par\vspace{2pt}
Output in the following format:\newline{}\texttt{\textasciigrave{}\textasciigrave{}\textasciigrave{}python}\newline{}\{\newline{}    "reasoning": "...",\newline{}    "action": "...",\newline{}\}\newline{}\texttt{\textasciigrave{}\textasciigrave{}\textasciigrave{}} \\[3pt]
\midrule
\textbf{Multiple-candidate prediction} \\[2pt]
\{grounding\}
\par\vspace{2pt}
Based on the features, infer \{n\_candidate\} diverse potential next actions (A concise sentence, \{character\}: ...) of \{character\} in the following scene:
\par\vspace{2pt}
\{scene\}
\par\vspace{2pt}
Output in the following format (same language as the given scene):\newline{}\# Reasoning\newline{}...\newline{}\# Actions\newline{}1. ...\newline{}2. ...\newline{}... \\[3pt]
\bottomrule
\end{tabular}
\end{table}

\begin{table}[htbp]
\caption{NAP evaluation prompt and dimension-specific criteria.}
\label{tab:evaluation_prompts}
\centering
\footnotesize
\setlength{\tabcolsep}{0pt}
\begin{tabular}{@{}>{\raggedright\arraybackslash}p{\linewidth}@{}}
\toprule
\textbf{Common evaluation template} \\[2pt]
\# Scene\newline{}\{scene\}
\par\vspace{2pt}
Your Response: \{prediction\}\newline{}Ground Truth: \{action\}
\par\vspace{2pt}
\{dimension\_instruction\}
\par\vspace{2pt}
Output only one lowercase word: yes or no. \\[3pt]
\midrule
\textbf{Emotion} \\[2pt]
Decide whether the emotion expressed by \{character\} in "Your Response" matches the emotion expressed by \{character\} in "Ground Truth".
\par\vspace{2pt}
Judge only the character's dominant expressed emotion, such as fear, anger, surprise, joy, sadness, relief, embarrassment, or calmness. Minor differences in intensity or wording do not matter. Ignore intent, causality, position, and semantic content.
\par\vspace{2pt}
If neither text expresses a clear emotion, treat them as matching. If only one text expresses a clear emotion, treat them as not matching. \\[3pt]
\midrule
\textbf{Intent} \\[2pt]
Decide whether the immediate intent of \{character\} in "Your Response" matches the immediate intent of \{character\} in "Ground Truth".
\par\vspace{2pt}
Intent means what the character is trying to accomplish through the utterance or action, such as obtaining information, informing someone, refusing, persuading, comforting, threatening, escaping, investigating, or ending an interaction. Judge the immediate purpose rather than the exact wording or information conveyed.
\par\vspace{2pt}
Ignore emotion, causality, position, and semantic content. \\[3pt]
\midrule
\textbf{Causality} \\[2pt]
Decide whether the character logic behind \{character\}'s utterance or action in "Your Response" matches the character logic behind it in "Ground Truth".
\par\vspace{2pt}
Character logic means the underlying reasoning that connects the character's understanding of the situation, beliefs, priorities, and disposition to the utterance or action. The exact immediate trigger does not need to be identical. Treat the two texts as matching when they follow the same underlying character reasoning, and as not matching when they rely on different or incompatible character reasoning.
\par\vspace{2pt}
Do not require emotion, intent, position, semantic content, or wording to match by themselves. Judge only whether the character logic is the same. \\[3pt]
\midrule
\textbf{Position} \\[2pt]
Decide whether the position (\promptpositiongloss{}) expressed by \{character\} in "Your Response" matches the position (\promptpositiongloss{}) expressed by \{character\} in "Ground Truth".
\par\vspace{2pt}
Position (\promptpositiongloss{}) means the character's own position in the situation: where the character stands and how the character aligns themself with what is happening, such as supporting or opposing, accepting or resisting, committed or hesitant, proactive or passive, serious or playful, or neutral. A position does not require a specific target or object. Judge the character's own \promptpositiongloss{}, not merely an attitude toward another person.
\par\vspace{2pt}
Ignore emotion, intent, causality, and semantic content. If neither text expresses a marked position, treat both as neutral and matching. \\[3pt]
\midrule
\textbf{Content} \\[2pt]
Decide whether the main content of \{character\}'s utterance or action in "Your Response" matches the main content in "Ground Truth".
\par\vspace{2pt}
Judge only the single most central proposition, question, answer, decision, or concrete action in each text. If both texts communicate or perform the same main point, treat them as matching even when secondary content, supporting details, modifiers, examples, or extra sub-actions differ. If multiple contents appear, compare the most central one in each text.
\par\vspace{2pt}
Ignore emotion, intent, causality, position, and wording style. Treat them as not matching only when the main content itself differs. \\[3pt]
\bottomrule
\end{tabular}
\end{table}

\clearpage
\section{Case Studies}
\label{apdx:case_studies}

In the Manosaba cases in Table~\ref{tab:mcq_case_studies}, active grounding recovers earlier observations that are decisive for the current question. The proposed sections bring back the spray-paint smell in one case and distinguish an oral wound from fingertip bloodstains in the other, supporting correct answers where Vanilla fails. This illustrates the value of organizing storyline memory around the current information need and retaining the distinctions required for grounded prediction.

\begin{table}[htbp]
\caption{MCQ cases from Manosaba. Earlier story provides reader-only background; Current scene and options are excerpts of the shared local input. Three actual state-bookmark proposals and their synchronized contents are shown per case, with the primary supporting entry in bold. Text is translated and abridged.}
\label{tab:mcq_case_studies}
\centering
\small
\setlength{\tabcolsep}{4.5pt}
\renewcommand{\arraystretch}{1.06}
\begin{tabularx}{\textwidth}{@{}>{\RaggedRight\arraybackslash}p{0.305\textwidth} >{\RaggedRight\arraybackslash}X >{\RaggedRight\arraybackslash}p{0.13\textwidth}@{}}
\toprule
\textbf{Context \& Question} & \textbf{Bookmarks} & \textbf{Answers} \\
\midrule
\textbf{Act 1, Chapter 1}\par\smallskip \emph{Earlier story.} Noa was found dead in her cell, and Ema investigated its bloodstained, painted floor and nearby objects. Coco and Milia had passed the cell after a livestream without noticing a body; Coco therefore argued that Noa was still alive then.\par\smallskip \emph{Current scene (trial).}\par \emph{Coco:} ``With that much blood, wouldn't we have noticed the smell?''\par \emph{Ema:} ``What could tell us whether there was a smell at the time?''\par\smallskip A. Butterfly painted on the floor; B. Photo of fruit; C. Spray paint; D. Disassembled parts.
 & \emph{Proposed sections:} Ema's recalled scene observations; \textbf{Material used for the butterfly drawing}; Origin and use of the disassembled parts.\par\medskip \emph{Ema's recalled scene observations:} During the trial, Ema noted scratches on the floor but no white paint on the arrow. She argued that another object had made the scratches, then continued examining when Noa died and whether someone could enter the cell.\par\medskip \textbf{Material used for the butterfly drawing: During the trial, Ema recalled the clues found at Noa's crime scene: a white-painted floor, a crossbow arrow, a butterfly drawn in blood, a spray-paint smell, scratches, and a bucket of mechanical parts.}\par\medskip \emph{Origin and use of the disassembled parts:} Ema and Meruru had found dismantled crossbow parts in a corridor bucket the previous evening. During the trial, Ema argued that the crossbow was already dismantled while Noa was alive and was difficult to reassemble.
 & Baseline:\par \textcolor{red!75!black}{B (incorrect)}\par\medskip Bookmarks:\par \textcolor{green!45!black}{C (correct)} \\
\midrule
\textbf{Act 1, Chapter 5}\par\smallskip \emph{Earlier story.} Nanoka's body was discovered as the execution platform rose from underground. It appeared partly frozen. Ema examined the body before the trial, but the cause of death remained unknown.\par\smallskip \emph{Current scene (trial).}\par \emph{Margo:} ``Nanoka's body had no wounds.''\par \emph{Ema:} ``Wait, there was a wound. It was ...''\par\smallskip A. Fingertips; B. Heart; C. Inside her mouth.
 & \emph{Proposed sections:} \textbf{Injury locations Ema observed on Nanoka's body}; Known information on Nanoka's cause of death; Bloodstains on Nanoka's fingertips and hands.\par\medskip \textbf{Injury locations Ema observed on Nanoka's body: On examining the body, Ema found a wound inside Nanoka's mouth and faint bloodstains on her fingertips. No other conspicuous external injuries were observed.}\par\medskip \emph{Known information on Nanoka's cause of death:} The cause of death remained undetermined. Recorded clues included faint fingertip bloodstains, a note marked 169231 in her pocket, and her gun alongside bullet marks in ice in the freezer room.\par\medskip \emph{Bloodstains on Nanoka's fingertips and hands:} At the time of the trial, the record described faint bloodstains adhering to Nanoka's fingertips, but no wounds on the fingertips themselves.
 & Baseline:\par \textcolor{red!75!black}{B (incorrect)}\par\medskip Bookmarks:\par \textcolor{green!45!black}{C (correct)} \\
\bottomrule
\end{tabularx}
\end{table}

\clearpage
\section{Evaluation Generality Validation}
\label{apdx:evaluation_generality}

\begin{figure}[htbp]
    \centering
    \includegraphics[width=0.75\linewidth]{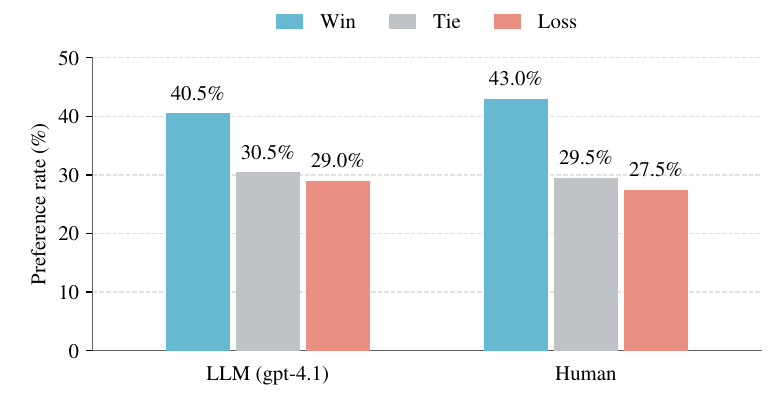}
    \caption{Open-domain role-playing evaluation against 3-way ETA on $200$ scenes. Win and loss denote preference for \our and 3-way ETA, respectively.}
    \label{fig:open_domain}
\end{figure}

\paragraph{Open-domain Test}
We evaluate \our on $200$ new starting scenes involving interactions between characters from the same artifact, with $10$ turns generated per scene. We compare \our against 3-way ETA using both \texttt{gpt-4.1} and human judges. Figure~\ref{fig:open_domain} reports the win, tie, and loss rates. Both judges prefer \our more often than 3-way ETA, supporting its generalization to unseen interactive scenarios.

\clearpage
\section{Artifact and Character Background}
\label{apdx:character_info}

Tables~\ref{tab:story_info} and~\ref{tab:character_info} summarize the six evaluated artifacts and their characters.

\begin{table}[htbp]
\caption{Background of the six evaluated artifacts.}
\label{tab:story_info}
\centering
\small
\begin{tabularx}{\linewidth}{@{}>{\raggedright\arraybackslash}p{0.28\linewidth}>{\raggedright\arraybackslash}X@{}}
\toprule
\textbf{Artifact} & \textbf{Background} \\
\midrule
\textit{The Melancholy of Haruhi Suzumiya} (Haruhi) &
A school comedy and mystery centered on Haruhi, whose impulses unknowingly reshape reality. Kyon and the SOS Brigade navigate ordinary club activities alongside supernatural incidents. \\
\midrule
\textit{JoJo's Bizarre Adventure} (JOJO) &
In Part 3, Jotaro and his companions journey across the world to confront DIO. Their encounters combine tactical Stand battles with flamboyant personalities and tense alliances. \\
\midrule
\textit{Fullmetal Alchemist} (FMA) &
Two brothers seek to restore their bodies after a forbidden alchemical ritual. Their journey explores sacrifice, political conflict, and the human cost of power. \\
\midrule
\textit{A Game of Thrones} (AGOT) &
Noble houses compete for power through alliances, betrayal, and war. Family loyalties and personal ambitions shape their choices amid growing political conflict. \\
\midrule
\textit{Poppin'Party} (PoPiPa) &
Five high school friends form a band and pursue their musical ambitions. Band Story 0 and Event 321 explore their relationships through practice, performance, and everyday life. \\
\midrule
\textit{Magical Girl Witch Trials} (Manosaba) &
Thirteen girls imprisoned as suspected witches face murders and trials. Their search for the truth depends on testimony, accumulated clues, and each girl's magical abilities. \\
\bottomrule
\end{tabularx}
\end{table}

\clearpage
\begingroup
\small
\begin{longtable}{@{}>{\raggedright\arraybackslash}p{0.12\linewidth}>{\raggedright\arraybackslash}p{0.14\linewidth}>{\raggedright\arraybackslash}p{\dimexpr0.74\linewidth-4\tabcolsep\relax}@{}}
\caption{Background of the evaluated characters.}\label{tab:character_info}\\
\toprule
\textbf{Artifact} & \textbf{Character} & \textbf{Background} \\
\midrule
\endfirsthead
\multicolumn{3}{c}{\tablename~\thetable{} (continued)} \\
\toprule
\textbf{Artifact} & \textbf{Character} & \textbf{Background} \\
\midrule
\endhead
\midrule
\multicolumn{3}{r}{Continued on next page} \\
\endfoot
\endlastfoot
Haruhi & Haruhi & An impulsive and hyperactive high school girl whose unusual worldview sets off many of the story's strange events. \\*
 & Kyon & A sarcastic and level-headed student who narrates the story while serving as Haruhi Suzumiya's reluctant stabilizing partner. \\*
 & Nagato & A quiet and unreadable SOS Brigade member with extraordinary intelligence and mysterious otherworldly origins. \\*
 & Koizumi & A smiling transfer student and esper who supports the Brigade while carefully concealing important secrets. \\*
 & Asahina & A timid and gentle upperclassman drawn into the SOS Brigade as its cute and enigmatic mascot, often swept into its chaos. \\
\midrule
JOJO & Jotaro & A stoic high schooler and leader of the Stardust Crusaders, known for Star Platinum and unshakable resolve. \\*
 & Polnareff & A bold French swordsman whose flamboyant personality and Silver Chariot make him a vivid ally. \\*
 & Joseph & A quick-thinking Joestar whose tricks, exaggeration, and improvisation often turn battles in his favor. \\*
 & DIO & A charismatic and ruthless vampire whose ambition and overwhelming presence define him as a legendary enemy. \\*
 & Kakyoin & A calm and analytical Crusader who fights with Hierophant Green and its precise emerald attacks. \\*
 & Avdol & A wise and loyal Egyptian Stand user whose Magician's Red controls powerful flames. \\*
 & Iggy & A stubborn Boston Terrier Stand user who dislikes cooperation but proves unexpectedly important. \\
\midrule
FMA & Edward & A talented and stubborn young alchemist traveling to restore the bodies he and his brother lost. \\*
 & Alphonse & A kind boy whose soul is bound to a suit of armor, journeying with his brother to recover what was taken. \\*
 & Winry & A gifted automail mechanic and childhood friend of the Elrics, known for technical skill and steady compassion. \\*
 & Roy & A charismatic State Alchemist and flame user who seeks to reform the military from within. \\*
 & Ling & A determined prince from Xing who pursues immortality while carrying the burden of national duty. \\
\midrule
AGOT & Tyrion & The sharp-witted youngest Lannister, navigating politics with intelligence, courage, and dark humor despite lifelong scorn. \\*
 & Daenerys & An exiled Targaryen princess who grows from a hesitant political pawn into a determined claimant to power. \\*
 & Cersei & An ambitious queen whose beauty masks ruthless devotion to family, status, and control. \\*
 & Jaime & The infamous Kingslayer, charming and deadly yet deeply conflicted by duty, loyalty, and reputation. \\*
 & Robb & The dutiful heir of Winterfell, forced into leadership and war before he is ready. \\*
 & Eddard & The honorable Lord of Winterfell, stern in duty and deeply committed to justice and family. \\*
 & Arya & An independent Stark girl who rejects courtly expectations in favor of freedom, survival, and the blade. \\*
 & Catelyn & The practical Lady of Winterfell, driven by fierce maternal loyalty and political urgency. \\*
 & Sansa & The elder Stark daughter, raised with courtly dreams that are tested by harsh political reality. \\*
 & Jon & Eddard's brooding illegitimate son, shaped by questions of identity, duty, and belonging. \\*
 & Bran & A curious young Stark whose fall places him on an unexpected and consequential path. \\
\midrule
PoPiPa & Kasumi & An upbeat vocalist and guitarist whose starry enthusiasm pulls others forward and begins the band's journey. \\*
 & Tae & A free-spirited lead guitarist with strong technique and quirky instincts, often moving at her own pace while enriching the band's sound. \\*
 & Rimi & A shy and gentle bassist who gains confidence through performance and supports the group with warmth. \\*
 & Saaya & A caring and dependable drummer whose family-minded steadiness anchors the band in practice and daily life. \\*
 & Arisa & A sharp-tongued but reliable keyboardist whose practicality and quick judgment keep the band organized and grounded. \\
\midrule
Manosaba & Gokucho & The mysterious owl warden who oversees the prison, its inmates, and the witch trials. \\*
 & Ema & A warm, outgoing girl who longs for friendship and fears rejection. \\*
 & Hiro & A diligent, principled student whose strong sense of justice leaves little room for compromise. \\*
 & An-An & A withdrawn girl who rarely speaks and communicates through a sketchbook. \\*
 & Noah & An aloof street artist whose celebrated work conceals its creator's identity. \\*
 & Leia & A popular actress who feels responsible for leading and protecting others. \\*
 & Miria & A gentle, timid girl whose flashy appearance contrasts with her hesitant manner. \\*
 & Margo & An affable con artist who keeps others at a distance through concealed distrust. \\*
 & Nanoka & A guarded loner who knows the prison well and resents its authority. \\*
 & Alisa & A confrontational runaway who dislikes others and, most of all, herself. \\*
 & Sherry & A cheerful girl who calls herself a detective and has no moral scruples. \\*
 & Hanna & A proud girl who poses as an aristocrat despite growing up poor. \\*
 & Coco & A streamer whose cheerful persona conceals a sharp tongue and volatile moods. \\*
 & Meruru & An anxious, sensitive girl who worries about others and doubts herself. \\
\bottomrule
\end{longtable}
\endgroup


\end{document}